\def\year{2020}\relax
\documentclass[letterpaper]{article} 
\usepackage{aaai20}  
\usepackage{times}  
\usepackage{helvet} 
\usepackage{courier}  
\usepackage[hyphens]{url}  
\usepackage{graphicx} 

\usepackage{times}  
\usepackage{helvet}  
\usepackage{courier}  
\usepackage{multirow}
\usepackage{amsmath}
\usepackage{bm}

\usepackage{graphicx}  
\title{Fine-Tuning BERT for Schema-Guided Zero-Shot Dialogue State Tracking}

\author{Yu-Ping Ruan, Zhen-Hua Ling, Jia-Chen Gu, Quan Liu\\ 
National Engineering Laboratory for Speech and Language Information Processing,\\
University of Science and Technology of China, Hefei, P.R.China\\
\tt \{ypruan, jcgu\}@mail.ustc.edu.cn, \{zhling, quanliu\}@ustc.edu.cn 
}
 
\begin{document}

\maketitle

\begin{abstract}
We present our work on Track 4 in the Dialogue System Technology Challenges 8 (DSTC8). The DSTC8-Track 4 aims to perform dialogue state tracking (DST) under the zero-shot settings, in which the model needs to generalize on unseen service APIs given a schema definition of these target APIs. Serving as the core for many virtual assistants such as Siri, Alexa, and Google Assistant, the DST keeps track of the user's goal and what happened in the dialogue history, mainly including intent prediction, slot filling, and user state tracking, which tests models' ability of natural language understanding. Recently, the pretrained language models have achieved state-of-the-art results and shown impressive generalization ability on various NLP tasks, which provide a promising way to perform zero-shot learning for language understanding. Based on this, we propose a schema-guided paradigm for zero-shot dialogue state tracking (SGP-DST) by fine-tuning BERT, one of the most popular pretrained language models. The SGP-DST system contains four modules for intent prediction, slot prediction, slot transfer prediction, and user state summarizing respectively. According to the official evaluation results, our SGP-DST (team12) ranked 3rd on the joint goal accuracy (primary evaluation metric for ranking submissions) and 1st on the requsted slots F1 among 25 participant teams.
\end{abstract}

\section{Introduction} \label{sec:intro}
Virtual assistants have been commercialized and provide many services such as finding flights, checking weather, and booking hotels. Among these, the Siri, Alexa, Google Assistant, and Cortana are the most popular and advanced frameworks, which provide a conversational interface to a large number of services and APIs spanning multiple domains. Dialogue state tracking (DST) is a core component of such task-oriented dialogue systems. The DST keeps track of the user's goal and what happened in the dialogue history, and output the dialogue state after each user utterance, which is a summary of the entire conversation till the current turn. The dialogue state then be used to determine what action should be taken by the system in next steps.

Deep learning models have achieved state-of-the-art results in dialogue state tracking \cite{mrkvsic2017neural,liu2017end,rastogi2018multi,zhong2018global}. Common public datasets for DST like DSTC2 \cite{henderson2014second}, MultiWOZ \cite{budzianowski2018multiwoz}, and M2M \cite{shah2018building} cover few domains and assume a single static ontology per domain, which do not sufficiently capture a number of challenges that arise with scaling DST in production \cite{rastogi2019towards}. The DST need to support a large, and constantly increasing number of services over a large number of domains.

To highlight these challenges, the DSTC8-Track 4 \cite{rastogiscalable} presents the task of schema-guided zero-shot dialogue state tracking and released the schema-guided dialogue (SGD) dataset \cite{rastogi2019towards}, which is the largest public task-oriented dialogue corpus, with over $16,000$ dialogues in training set spanning 26 services belong to 16 domains. Specifically, the SGD dataset is designed to develop and test models' ability to generalize in zero-shot settings since the evaluation sets in SGD contain unseen services and domains.
Zero-shot DST models utilizing domain and/or slot descriptions have gaining popularity for spoken language understanding tasks \cite{bapna2017towards,kumar2017zero,lee2019zero}. The SGD dataset in DSTC8-Track 4 also aims to motivate similar approaches for dialogue state tracking \cite{rastogiscalable}.
The targets of DST on the SGD dataset mainly include the intent prediction, and slot filling.

The DST task relies on the models' ability of natural language understanding. Recently, the pretrained language models have achieved state-of-the-art results and shown impressive generalization ability on a wide variety of NLP tasks. Among these, the GPT \cite{radford2018improving}, ELMo \cite{peters2018deep}, BERT \cite{devlin2019bert}, and XLNet \cite{yang2019xlnet} are the most popular and advanced ones. These pretrained language models provide a promising way to perform zero-shot learning for language understanding.
Considering the zero-shot settings in the SGD dataset, we propose a schema-guided paradigm for zero-shot dialogue state tracking (SGP-DST) by fine-tuning BERT \cite{devlin2019bert}. The SGP-DST system contains four modules:
1) \emph{intent prediction}: which aims to predict the user intent for each user utterance,
2) \emph{slot prediction}: which aims to predict slots value and user requested slots from corresponding user utterance, 3) \emph{slot transfer prediction}: which aims to decide which slot value should be transferred (copied) from system history actions or user history states, and 4) \emph{user state summarization}: which aims to summarize the prediction results from previous 3 modules and update the user states for current turn. According to the official evaluation results\footnote{\url{https://docs.google.com/spreadsheets/d/19Z1e1mXch4HnPoXfMGxw2UEHBt3cTTcwulZ1H_7U7DM/edit#gid=1291434552}}, our proposed SGP-DST system (team12) ranked 3rd among 25 participant teams.

\section{The Schema-Guided Dialogue Dataset}
\begin{table}[!t]
	\centering
    \small
	\begin{tabular}{p{7.5cm}}
        \hline
        \hline
		\textbf{service\_name}: ``Banks\_1" \\
		\textbf{desctiption}: ``Manage bank accounts and transfer money" \\
        \hline
        \hline
        \textbf{slot\_name}: ``account\_type" \\
        \textbf{categorical}: True\\
        \textbf{desctiption}: ``The account type of the user" \\
        \textbf{possible\_values}: [``checking", ``savings"] \\
        \hline
        \textbf{slot\_name}: ``amount" \\
        \textbf{categorical}: False\\
        \textbf{desctiption}: ``The amount of money to transfer" \\
        \textbf{possible\_values}: [] \\
        \hline
        \hline
        \textbf{intent\_name}: ``CheckBalance" \\
        \textbf{is\_transactional}: False\\
        \textbf{desctiption}: ``Check the amount of money in a user's bank account" \\
        \textbf{required\_slots}: [``account\_type"] \\
        \textbf{optional\_slots}: [] \\
        \hline
        \hline
	\end{tabular}
    \caption{Example schema (incomplete) for a bank service.}
	\label{tab:schema}
\end{table}

\begin{table}[!t]
	\centering
	\begin{tabular}{c|ccc}
        \hline
        \hline
         & Train & Dev & Test \\
         \hline
         No. of dialogs & 16,142 & 2,482 & 4201\\
         No. of domains & 16 & 16 & 18 \\
         No. of services & 26 & 17 & 21 \\
         Avg. turns per dialogue & 20.44 & 19.63 & 20.13 \\
         Avg. tokens per turn & 9.75 & 9.66 & 10.40 \\
         dialogs with unseen APIs(\%) & 0.0 & 42.01 & 69.64 \\
         \hline
         \hline

	\end{tabular}
    \caption{Statistics of the SGD dataset. }
	\label{tab:sgd_statistics}
\end{table}

The schema-guided dialogue (SGD) dataset consists of dialogues between a human and a virtual assistant, which are generated with the help of a dialogue simulator and crowd-workers along with schemas for one or more APIs relevant to the dialogues \cite{rastogi2019towards}.
A schema defines the interface for a backend service API and contains a description, a set of slots and intents as shown in Table \ref{tab:schema}. The description is a natural language summary of the function of the API. Different from traditional schema definitions, the schemas in SGD also gives a natural language description for each slot and intent to help models generalize to unseen API schemas. There are two types of slots in SGD dataset: 1) \emph{categorical}: a slot taking one of a finite set of possible values, which have been included in corresponding schema definition. 2) \emph{free-form}: a slot can take any string value, which can be derived from the dialogue history.

The dialogue state tracking on SGD dataset contains following sub-targets: 1) predicting the user's intent for each user utterance, 2) predicting the slots value, which may be from corresponding user utterance or transferred (copied) from the system history actions or user history states, and 3) predicting which slots are requested by the user.

Some main statistics of the SGD dataset are shown in Table \ref{tab:sgd_statistics}.
Except for the number of total dialogues, service domains, and service APIs, Table  \ref{tab:sgd_statistics} also gives the average turns per dialogue, average tokens per turn, and percentage of dialogues with unseen service APIs. We can find that the test set has more dialogues than the dev set, and more critically, the test set includes more dialogues with unseen APIs than the dev set.
For more details about the SGD dataset, we recommend readers to check the descriptions on the data website\footnote{\url{https://github.com/google-research-datasets/dstc8-schema-guided-dialogue}}.

\section{System Description}
As introduced in Section \ref{sec:intro}, our proposed SGP-DST system has four modules, which work jointly in a pipeline manner to track the user dialogue state. The \emph{intent prediction}, \emph{slot prediction}, and \emph{slot transfer prediction} modules are all based on fine-tuned BERT, and the last \emph{user state summarizing} module just merges the results from previous three modules based on simple rules. Followings are the details about the module design.

\subsection{Intent Prediction}\label{sec:intent_pred}
\begin{figure}[!t]
	\centering
    \small
	\includegraphics[width=3.5in]{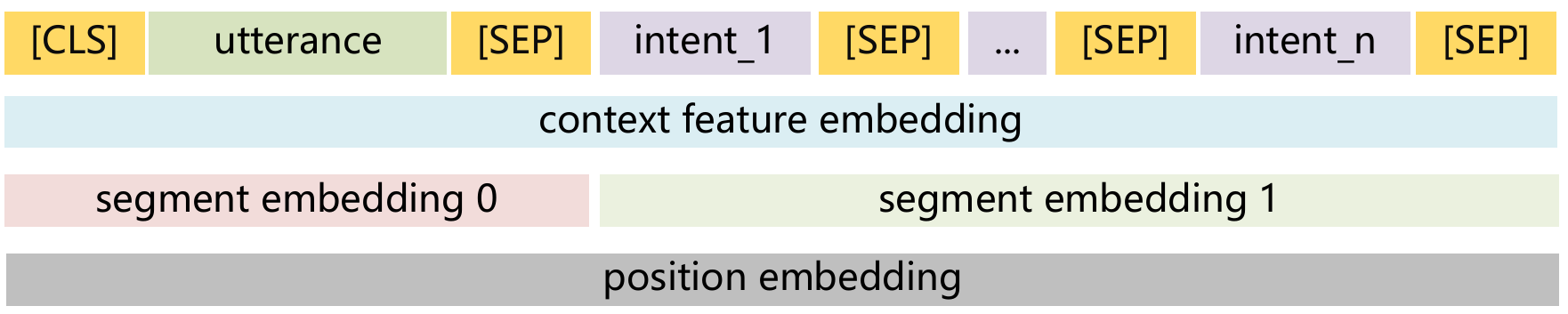}
    \caption{BERT input representation in the \emph{intent prediction} module. Except for the token embedding, segment embedding, and position embedding, we introduce another context feature embedding.}
	\label{fig:intent_pred}
\end{figure}

As shown in Figure \ref{fig:intent_pred}, the inputs to BERT are the sum of the token embeddings, segment embeddings, position embeddings, which are the same with those in original BERT \cite{devlin2019bert}, and additionally context feature embeddings. The token sequences are the concatenation of utterance tokens and intents description tokens. Specifically, we concatenate the last system utterance and the current user utterance in form of ``sys: [system\_utterance] usr: [user\_utterance]" as the input utterance, and we pad the intents description fragments to the max number of intents considering that different service APIs have different number of intents.

We design the context feature as the intent indicator of last user utterance, considering that the user intent prediction should be transited smoothly from last user intent. So there are only two kinds of context feature embeddings, which is set to ``1" when the token belongs to last intent description and else is ``0". We use the ground-truth last intent labels for training but predicted intent labels for inference when derive the intent indicator feature.

Different from the original attention flow in BERT, in which each token can attend to all tokens in the input sequence, the tokens in intent\_i can only attend to the utterance tokens and the tokens belong to the same intent\_i. For the tokens in the utterance, they can attend to all tokens in the input sequence.

After BERT output the encoded sequence, we derive a contextual representation vector $\mathbf{r}_i$ for intent\_i by max pooling the encoded token representations belongs to correspond intent description. Then the representation vector $\mathbf{r}_i$ will be input to a feed-forward network as follows,
\begin{align}
s_i &= \mathbf{v}^T tanh(W \mathbf{r}_i + \mathbf{b}) \\
{prob}_i &= softmax(s_i)
\end{align}
in which the ${prob}_i$ gives the probability of selecting intent\_i as the user intent.
The cross-entropy between the predicted and ground truth distribution of intents is defined as the loss for training.

\subsection{Slot Prediction}
The \emph{slot prediction} module has three submodules for categorical slot value prediction, free-form slot value prediction, and requested slot prediction respectively. Table \ref{fig:slot_pred} shows an example for the slot prediction targets.

\begin{table}[!t]
	\centering
    \small
	\begin{tabular}{p{7.5cm}}
        \hline
        \hline
        \textbf{Service\_name}: ``Restanrants\_2" \\
        \hline
        \textbf{user utterance}: ``I want to make the reservation for \textbf{2 people} at \textbf{half past 11 in the morning}, and can you give me the \textbf{address}?"\\
        \hline
        \textbf{categorical slots}: [``number\_of\_seats": ``2"]\\
        \textbf{free-form slots}: [``time": ``half past 11 in the morning"]\\
        \textbf{required slots}: [``address"]\\
        \hline
        \hline
	\end{tabular}
    \caption{Example of slot prediction targets, including categorical slot value prediction, free-form slot value prediction, and required slot prediction.}
	\label{tab:slot_prediction}
\end{table}

\subsubsection{Categorical slot value prediction} 
\begin{figure}[!t]
	\centering
    \small
	\includegraphics[width=3.5in]{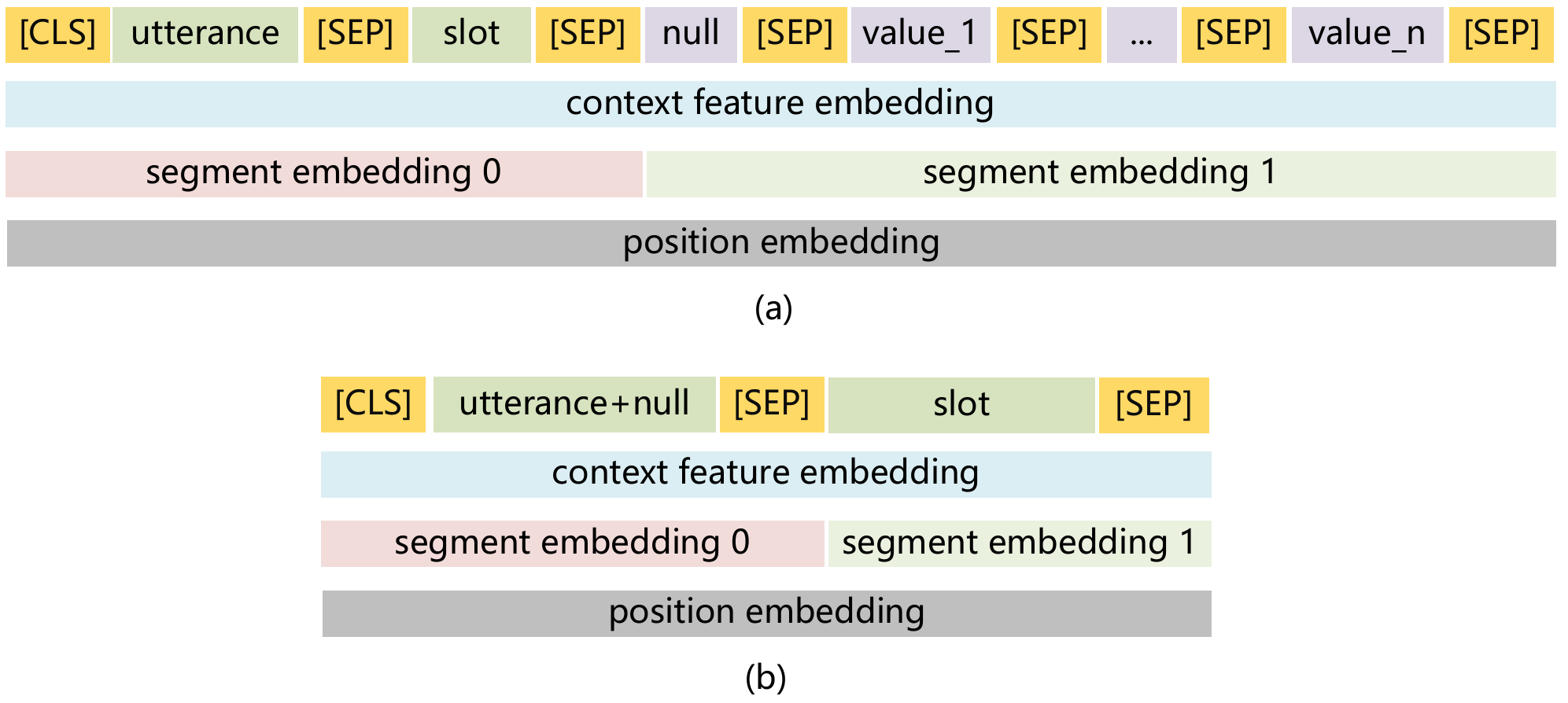}
    \caption{BERT input representation in the \emph{slot prediction} module: a) for categorial slot value prediction, and b) for free-form slot value prediction or requested slot prediction.}
	\label{fig:slot_pred}
\end{figure}

The BERT input representation for categorical slot value prediction is presented in Figure \ref{fig:slot_pred}(a). The input tokens contain the utterance tokens, slot description tokens, and corresponding possible values tokens, in which the utterance tokens are the concatenation of the last system utterance and the current user utterance. The slot description tokens are the concatenation of slot name and original slot description tokens. The possible value description fragments are padded into the max number of possible values. Additionally, we add the ``null" value as there may be no corresponding value present in the utterance.

The context feature here gives the information about whether the slot has been requested by system and whether system has offered an value for current slot, which are both binary. So there is totally 4 kinds of context feature embeddings. We derive the context feature from history system actions.

For the attention flow, the tokens in the utterance and the slot description can attend to all tokens in the input sequence. Similar with the attention flow of the intent tokens in Section \ref{sec:intent_pred}, the tokens in value\_i can only attend to the tokens in utterance, slot description, and tokens belong to value\_i.

For the final categorical slot value prediction, we adopt the same manner for intent prediction in Section \ref{sec:intent_pred}: first derive a contextual representation vector for each possible value including the ``null" value, then feed these contextual representation vectors into a feed-forward network with softmax output to get the probability distribution along all possible values.
The training target is the cross-entropy between the predicted and ground truth distribution of categorical slot values.

\subsubsection{Free-form slot value prediction} We build the free-form slot value prediction as a reading comprehension task, in which the model need to return a text span from the user utterance as the predicted slot value.
The input representation for BERT is shown in Figure \ref{fig:slot_pred}(b), in which the utterance tokens, slot description tokens, and context feature are derived with the same manner for categorical slot value prediction. Specifically, the utterance tokens are further concatenated with the ``null" token, considering that there may be no value present in the utterance for corresponding slot. The attention flow is identical with that in original BERT.

As for the text span prediction, we adopt the same processing way of BERT on SQuAD task \cite{devlin2019bert}. Let the encoded representation vector for token $i$ in utterance as $\mathbf{t}_i$, then the probability of token $i$ being the start/end of the value span is computed as a dot product between $\mathbf{t}_i$ and start vector $\mathbf{s}$ or end vector $\mathbf{e}$ followed by a softmax:
\begin{align}
p_i^s &= \frac{e^{\mathbf{s}^T \mathbf{t}_i}}{\sum_{j} e^{\mathbf{s}^T \mathbf{t}_i}} \\
p_i^e &= \frac{e^{\mathbf{e}^T \mathbf{t}_i}}{\sum_{j} e^{\mathbf{e}^T \mathbf{t}_i}}
\end{align}

Then the maximum scoring span is used as the predicted slot value. And when the maximum scoring span is ``null", we do not return any predict results for the current concerning slot. The cross-entropy between the predicted and ground truth distribution of the start and end positions are used as the training objective.

\subsubsection{Requested slot prediction} The input representation for requested slot prediction is identical with that for free-form slot value prediction. We treat the requested slot prediction as a sequence-level classification task, and we adopt the same processing way as the one for sequence classification in \cite{devlin2019bert}, in which the final encoded vector for ``[CLS]" is fed into a classifier.

\subsection{Slot Transfer Prediction}
\begin{table}[!t]
	\centering
    \small
	\begin{tabular}{p{7.5cm}}
        \hline
        \hline
        \textbf{Service\_name}: ``Buses\_1" \\
        \hline
        \textbf{user}: ``Can you find a bus? It's for a group of 4."\\
        \textbf{system}: ``How about a bus with 0 stops, \textbf{departing at 7 am}, and costs \$29?" \\
		\textbf{user}: ``Okay, what bus station is it leaving from? What bus station am I arriving at?" \\
		\textbf{system}: ``The destination station is Santa Fe Depot and you will be departing from Downtown Station." \\
        \textbf{user}: ``\textbf{Sounds great. Book the bus.}"\\
        \hline
        \hline
	\end{tabular}
    \caption{Example of in-domain slot transfer. When user makes the decision by saying ``Sounds great. Book the bus.", the value of slot ``leaving\_time" should be copied from system history action ``OFFER: leaving\_time: 7 am".}
	\label{tab:in_domain_transfer}
\end{table}

\begin{table}[!t]
	\centering
    \small
	\begin{tabular}{p{7.5cm}}
        \hline
        \hline
        \textbf{Last turn}:\\
        \textbf{Service\_name}: ``RentalCars\_1" \\
        \textbf{System utterance:}: ``Your car has been reserved in NYC."\\
        \textbf{user utterance}: ``That's good." \\
        \textbf{user states}:\{{``pickup\_date"}: ``March 11th", ``pickup\_city": ``NYC",...\} \\
        \hline
        \textbf{Current turn}:\\
        \textbf{Service\_name}: ``Buses\_1" \\
        \textbf{System utterance:}: ``Would you want to get there by taxi?"\\
        \textbf{user utterance}: ``No. I'd like a bus to get there." \\
        \textbf{user states}:\{{``to\_location"}: ``NYC"\} \\
        \hline
        \hline
	\end{tabular}
    \caption{Example of cross-domain slot transfer. When the user state tracking switched from Service ``RentalCars\_1" to service ``Buses\_1", the value of target slot ``to\_location" in `Buses\_1" frame should be copied from the source slot ``pickup\_city" in ``RentalCars\_1" frame. Note that the source slot can also be from system history actions in ``RentalCars\_1" frame.}
	\label{tab:cross_domain_transfer}
\end{table}

The slot-value in user state may be transferred (copied) from corresponding slot-value in system history actions or user history states, which cannot be figured out from current user utterance. For the SGD dataset, which mainly contains multi-domain dialogues, the slot-value can be transferred in domain or cross domain, noted as ``in-domain slot transfer" and ``{cross-domain slot transfer}" here.

For the in-domain slot transfer, as shown in Table \ref{tab:in_domain_transfer}, when user makes an agreement or transaction with the system, some certain slots should be transferred from system history actions which provide the values for corresponding slots.
The table \ref{tab:cross_domain_transfer} shows one example for the cross-domain slot transfer, which usually happens when the user state tracking switches from one service API to another.

For both in-domain and cross-domain slot transfer, there may be multiple values for a slot in the dialogue history, we use the most recently mentioned slot value for transfer.

\begin{figure}[!t]
	\centering
    \small
	\includegraphics[width=3.2in]{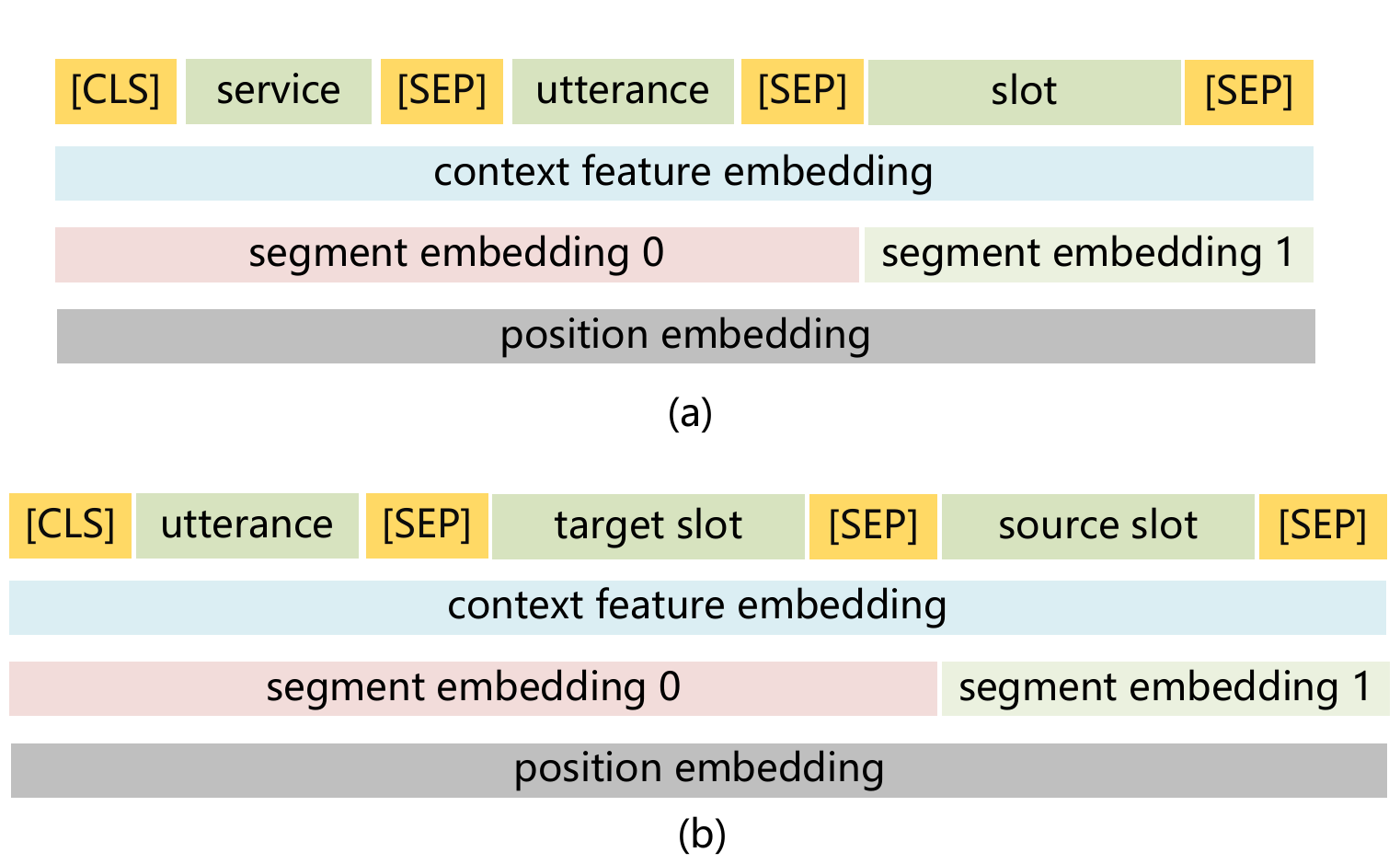}
    \caption{BERT input representation in the \emph{slot transfer prediction} module: a) for in-domain slot transfer prediction, and b) for cross-domain slot transfer prediction.}
	\label{fig:slot_transfer}
\end{figure}
\subsubsection{In-domain slot transfer}
Similar with previous settings, the BERT input representation for in-domain slot transfer is presented in Figure \ref{fig:slot_transfer}(a). The input tokens are the concatenation of the service description tokens, the utterance tokens, and the slot description tokens, in which the utterance and slot description tokens are derived with the same manner in categorical slot value prediction. Specifically, the context feature used here gives the information about whether the slot is optional/required in current user intent, whether system has given the value for current slot, and wether the slot has appeared in user history states, which contains four binary features. So there is totally 16 kinds of context feature embeddings. For the attention flow, it's identical with that in original BERT.

Finally, We treat the slot transfer prediction as a sequence-level classification task, and we adopt the standard processing way of BERT for sequence classification \cite{devlin2019bert}.

\subsubsection{Cross-domain slot transfer}
The BERT input representation for cross-domain slot transfer prediction is shown in Figure \ref{fig:slot_transfer}(b), in which the input tokens are the concatenation of utterance tokens, target slot tokens and source slot tokens. For the context feature, totally 5 binary features are used, i.e., whether current service frame is the continuation of previous turn, whether the target slot is optional/required in current user intent, whether the target slot has appeared in the history of current service frame, and whether the source slot has been in source service frame.
So we set total 32 kinds of context feature embeddings here.

The cross-domain slot transfer prediction is also treated as a sequence-level classification task, and the standard processing manner of BERT for sequence classification \cite{devlin2019bert} is adopted.

\begin{table*}[!ht]
	\centering
	\begin{tabular}{c|cccc}
        \hline
        \hline
        & \textbf{Active Intent Acc.} & \textbf{Requested Slot F1} & \textbf{Average Goal Acc.} & \textbf{Joint Goal Acc.} \\
        \hline
        & \multicolumn{4}{c}{\multirow{1}{*}{\emph{SGP-DST}}}\\
        \hline
        All APIs & 0.9529 & 0.9839 & 0.9387 & 0.8001 \\
        Seen APIs & 0.9571 & 0.9845 & 0.9659 & 0.8831 \\
        Unseen APIs & 0.9476 & 0.9832 & 0.9027 & 0.6923 \\
        \hline
        & \multicolumn{4}{c}{\multirow{1}{*}{\emph{SGP-DST without in-domain slot transfer}}}\\
        \hline
        All APIs & 0.9529 & 0.9839 & 0.8100 & 0.4975  \\
        Seen APIs & 0.9571 & 0.9845 & 0.8362 & 0.5416 \\
        Unseen APIs & 0.9476 & 0.9832 & 0.7753 & 0.4400 \\
        \hline
        & \multicolumn{4}{c}{\multirow{1}{*}{\emph{SGP-DST without cross-domain slot transfer}}}\\
        \hline
        All APIs & 0.9529 & 0.9839 & 0.8406 & 0.6361 \\
        Seen APIs & 0.9571 & 0.9845 & 0.8748 & 0.7106 \\
        Unseen APIs & 0.9476 & 0.9832 & 0.7954 & 0.5393 \\
        \hline
        & \multicolumn{4}{c}{\multirow{1}{*}{\emph{SGP-DST without in\&cross-domain slot transfer}}}\\
        \hline
        All APIs & 0.9529 & 0.9839 & 0.7048 & 0.3940 \\
        Seen APIs & 0.9571 & 0.9845 & 0.7353 & 0.4170 \\
        Unseen APIs & 0.9476 & 0.9832 & 0.6644 & 0.3640 \\
        \hline
        \hline

	\end{tabular}
    \caption{Evaluation results on the dev set. The ``all APIs", ``seen APIs", ``and unseen APIs" represents the whole dev set, subset whose APIs appear in traning set, and subset whose APIs are unseen in training set.}
	\label{tab:overall_results}
\end{table*}

\subsection{User State Summarization}
The \emph{user state summarization} module works at inference stage.
This module aims to summarize the prediction results from above three modules and update the user state at each turn.
Specifically, the \emph{intent prediction} and \emph{slot prediction} modules should first finish their corresponding prediction targets, then the \emph{slot transfer prediction} module can work for its targets since the user intent and user history states feature are used in this module.

The rules for user state summarizing and updating are very simple and mainly include: 1) for each prediction target in above three modules, we set a scoring threshold and only consider the predicted items whose accompanied prediction score returned by the model is above the threshold, in which the predicted probabilities are uses as the scores directly and the score thresholds are set to 0.8, 0.5, 0.9, 0.85, and 0.9 for categorical slot, free-form slot, requested slot, in-domain slot transfer, and cross-domain slot transfer prediction respectively. 2) we manually remove the slots which are not required or optional in corresponding user intent according the schema definition.

\section{Experiments}
\subsection{Training Details}
We extract training and dev samples from the original released SGD dialogue corpus for each module in our SGP-DST system.
All dialogues, including the single-domain and the multi-domain, in the SGD training set are used for the model training. We use the PyTorch implementation of BERT-base with the pretrained model file \texttt{bert-base-cased} provided by Google\footnote{\url{https://github.com/huggingface/pytorch-pretrained-BERT\#Fine-tuning-with-BERT-running-the-examples}}. For the fine-tuning process, we use mostly default settings. Specifically, the learning rate is 2e-05, the batch size is set to 128, and the max training epochs is 3. Totally, we fine-tuned 6 individual BERT-base models for intent, categorical slot, free-form slot, requested slot, in-domain slot transfer, and cross-domain slot transfer prediction respectively in our SGP-DST systems.

\subsection{Evaluation Metrics}
The official metrics for the evaluation of DST on the SGD dataset are listed below, in which the joint goal accuracy is used as the primary metric for ranking systems.

\begin{itemize}
\item \textbf{Active intent accuracy}: The fraction of user turns for which the active intent has been correctly predicted.
\item \textbf{Requested slots F1}: The macro-averaged F1 score for requested slots over the turns. For a turn, if there are no requested slots in both the ground truth and the prediction, that turn is skipped. The reported number is the average F1 score for all un-skipped user turns.
\item \textbf{Average goal accuracy}: This is the average accuracy of predicting the value of a slot correctly. A fuzzy matching based score is used for non-categorical slots. The slots which have a non-empty assignment in the ground truth dialogue state are only considered.
\item \textbf{Joint goal accuracy}: This is the average accuracy of predicting all slot assignments for a turn correctly. For non-categorical slots a fuzzy matching score is used to reward partial matches with the ground truth.
\end{itemize}

\begin{table*}[!t]
	\centering
	\begin{tabular}{c|cccc}
        \hline
        \hline
        & \textbf{Active Intent Acc.} & \textbf{Requested Slot F1} & \textbf{Average Goal Acc.} & \textbf{Joint Goal Acc.} \\
        \hline
        & \multicolumn{4}{c}{\multirow{1}{*}{\emph{With 0\% of dev set for training}}}\\
        \hline
        All APIs & 0.9529 & 0.9839 & 0.9387 & 0.8001 \\
        Seen APIs & 0.9571 & 0.9845 & 0.9659 & 0.8831 \\
        Unseen APIs & 0.9476 & 0.9832 & 0.9027 & 0.6923 \\
        \hline
        & \multicolumn{4}{c}{\multirow{1}{*}{\emph{With 50\% of dev set for training}}}\\
        \hline
        All APIs & 0.9620 & 0.9839 & 0.9557 & 0.8452  \\
        Seen APIs & 0.9625 & 0.9852 & 0.9755 & 0.9127  \\
        Unseen APIs & 0.9614 & 0.9823 & 0.9295 & 0.7573 \\
        \hline
        & \multicolumn{4}{c}{\multirow{1}{*}{\emph{With 90\% of dev set for training}}}\\
        \hline
        All APIs & 0.9629 & 0.9856 & 0.9631 & 0.8698 \\
        Seen APIs & 0.9633 & 0.9942 & 0.9794 & 0.9225 \\
        Unseen APIs & 0.9623 & 0.9864 & 0.9431 & 0.8059  \\
        \hline
        \hline
	\end{tabular}
    \caption{Evaluation results under the few-shot settings. We include 0\%, 50\%, and 90\% dialogues in dev set for training respectively and conduct the evaluation on the remaining dialogues in dev set.}
	\label{tab:few_shot}
\end{table*}

\begin{table*}[!t]
	\centering
	\begin{tabular}{c|cccc}
        \hline
        \hline
        & \textbf{Active Intent Acc.} & \textbf{Requested Slot F1} & \textbf{Average Goal Acc.} & \textbf{Joint Goal Acc.} \\
        \hline
        & \multicolumn{4}{c}{\multirow{1}{*}{\emph{With 0\% of dev set for training}}}\\
        \hline
        All APIs & 0.9185 & 0.9899 & 0.9125 & 0.7222 \\
        Seen APIs & 0.9564 & 0.9935 & 0.9582 & 0.8798 \\
        Unseen APIs & 0.9059 & 0.9887 & 0.8966 & 0.6696  \\
        \hline
        & \multicolumn{4}{c}{\multirow{1}{*}{\emph{With 90\% of dev set for training}}}\\
        \hline
        All APIs & 0.9234 & 0.9948 & 0.9199 & {0.7375}  \\
        Seen APIs & 0.9581 & 0.9965 & 0.9566 & 0.8795  \\
        Unseen APIs & 0.9118 & 0.9943 & 0.9071 & 0.6901 \\
        \hline
        & \multicolumn{4}{c}{\multirow{1}{*}{\emph{With 100\% of dev set for training}}}\\
        \hline
        All APIs & 0.9260 & 0.9954 & 0.9151 & {0.7257}  \\
        Seen APIs & 0.9574 & 0.9973 & 0.9631 & 0.8877  \\
        Unseen APIs & 0.9156 & 0.9947 & 0.8983 & 0.6717 \\
        \hline
        & \multicolumn{4}{c}{\multirow{1}{*}{\emph{Best submitted results among all participants}}}\\
        \hline
        Best Reported & 0.9692 & 0.9954 & 0.9712 & {0.8653}  \\
        Our Best (Rank) & \textbf{0.9260} (11) & \textbf{0.9954} (1) & \textbf{0.9199} (5) & \textbf{0.7375} (3)  \\
        \hline
        \hline
	\end{tabular}
    \caption{Official evaluation results on test set. We submitted the SGP-DST systems trained without dev samples and with 90\% dev samples respectively.}
	\label{tab:eval_on_test}
\end{table*}

\subsection{Results}

\subsubsection{Overall performance}
Table \ref{tab:overall_results} presents the overall performance of our SGP-DST system on the dev set. To analyze the SGP-DST's ability of zero-shot dialogue state tracking more comprehensively, we present the results on the whole dev set, subset whose APIs appear in training set, and subset whose APIs are unseen in training set, denoted as ``all APIs", ``seen APIs", ``and unseen APIs" respectively in Table \ref{tab:overall_results}.

We can find that there is a very small performance gap on the active intent and requested slot prediction between the ``seen APIs" subset and  the``unseen APIs" subset, which indicates that our SGP-DST system can generalize well on the active intent and requested slot prediction. However, on the metrics of average goal accuracy and joint goal accuracy, our SGP-DST system performs much worse on the ``unseen APIs" subset than on the ``seen APIs", which indicates that it's more difficult to generalize on the slot value prediction.

We also evaluate the importance of \emph{slot transfer prediction} module in our SGP-DST by removing the in-domain slot transfer prediction and cross-domain slot prediction respectively. We can find that both in-domain and cross-domain slot transfer has a significant effect on improving the system performance, especially for the in-domain slot transfer.

\subsubsection{Evaluation under few-shot settings} From results in Table \ref{tab:overall_results}, we can know that our system has significant poorer performance on the ``unseen APIs" subset than that on ``seen APIs" subset. Generally, zero-shot settings is a tough testing bed for neural network models, and it's actually an economy way to improve the models' performance by including a small number of unseen labelled samples into the training process, which is called few-shot learning. In order to do this, we simply random sample some dialogues in the dev set and adding them into the training process, then we evaluate the model on the remaining dialogues in the dev set.

Table \ref{tab:few_shot} presents the evaluation results of including dev samples for training. We can find that including samples from dev set for training can have a significant performance improvement on the joint goal prediction, especially for the ``unseen APIs" subset, which demonstrates the effects of few-shot learning.

\subsubsection{Official evaluation on test set} We submitted our SGP-DST systems trained without dev samples and with 90\%, and 100\% dev samples respectively for the final official evaluation on test set, whose results are present in Table \ref{tab:eval_on_test}.

For the SGP-DST system trained without dev samples, we can find that its performance on the ``seen APIs" test subset is very close to that on the ``seen APIs" dev subset, however, the performance on the ``unseen APIs" test subset is obviously poorer than that on the dev subset. Also, the SGP-DST system performs much worse on the whole test set than on the dev set since there is much more dialogues with unseen APIs in the test set.

For the SGP-DST systems trained with dev samples, both of them outperform the one trained without dev samples on all four metrics for the ``all APIs" set. Specifically, the system trained with extra 100\% dev samples achieves best results on the active intent accuracy and requested slot F1, however, the one trained with 90\% dev samples achieves best results on the average goal accuracy and joint goal accuracy, which means there may be overfitting for our SGP-DST systems trained with dev samples.

Compared with the bested reported results among all 25 participants, our SGP-DST system ranks 1st on the request slot F1, but achieves significantly worse result on the active intent prediction. For the slot value prediction, there is significant performance gap between our system and the best reported, our SGP-DST ranks 5th and 3rd on the average goal accuracy and joint goal accuracy respectively.

\section{Conclusion}
In this paper, we present our SGP-DST system for DSTC8-Track 4, which aims to perform dialogue state training (DST) under the zero-shot settings. Our proposed SGP-DST system includes four modules for intent prediction, slot prediction, slot transfer prediction, and user state summarizing respectively.
According to the official evaluation results, our SGP-DST (team12) ranked 3rd (primary evaluation metric for ranking submissions) and 1st on the requsted slots F1 among all 25 participant teams.
For the zero-shot dialogue state tracking, it's still worth our study and exploration for more powerful models with better generalization ability of language understanding.

\bibliography{dstc8}
\bibliographystyle{aaai}

\end{document}